%% file: main.tex
\documentclass{article}
\usepackage{ijcai24}

\usepackage{times}
\usepackage{soul}
\usepackage{url}
\usepackage[colorlinks=true,hyperfootnotes=false]{hyperref}
\usepackage[utf8]{inputenc}
\usepackage[small]{caption}
\usepackage{graphicx}
\usepackage{amsmath}
\usepackage{amsthm}
\usepackage{booktabs}
\usepackage{algorithm}
\usepackage{algorithmic}
\usepackage[switch]{lineno}
\usepackage{subcaption}

\usepackage{natbib}
\usepackage[T1]{fontenc}
\usepackage{textgreek}
\usepackage{amsfonts}
\usepackage{color}
\usepackage[dvipsnames]{xcolor}
\usepackage{longtable}
\usepackage{tabularx}
\usepackage{multirow}
\usepackage{bm}
\usepackage{enumitem}
\usepackage{listings} % Add this in your preamble
\usepackage{fancyvrb}
\usepackage[most]{tcolorbox}
\newcommand{\unnumberedfootnote}[1]{%
    {\let\thefootnote\relax\footnotetext{#1}}%
}

\title{REALM: RAG-Driven Enhancement of Multimodal Electronic Health Records Analysis via Large Language Models}

\author{
Yinghao Zhu$^{1~\ast}$\and
Changyu Ren$^{2~\ast}$\and
Shiyun Xie$^{1}$\and
Shukai Liu$^{2}$\and
Hangyuan Ji$^{2}$\and \\
Zixiang Wang$^{3}$\and
Tao Sun$^{2}$\and
Long He$^{1}$\and
Zhoujun Li$^{2}$\and
Xi Zhu$^{4}$\and
Chengwei Pan$^{1~\dagger}$\\
\affiliations
$^1$Institute of Artificial Intelligence, Beihang University, Beijing, China\\
$^2$School of Computer Science and Engineering, Beihang University, Beijing, China\\\
$^3$School of Software, Beihang University, Beijing, China\\\
$^4$China Mobile Research Institute, Beijing, China\\\
\emails
zhuyinghao@buaa.edu.cn, pancw@buaa.edu.cn
}

\begin{document}
\maketitle
\unnumberedfootnote{$^\ast$ Equal contribution, $^\dagger$ Corresponding author.}

\input{sections/abstract}

\input{sections/1.intro}
\input{sections/2.related_work}
\input{sections/3.problem_formulation}
\input{sections/4.methods}
\input{sections/5.experimental_setups}

\input{sections/6.experimental_results}

\input{sections/7.discussions}
\input{sections/8.conclusions}

\section*{Ethical Statement}

This study, involving the analysis of Electronic Health Records (EHR) using the MIMIC dataset, is committed to upholding high ethical standards. The MIMIC dataset is a de-identified dataset, ensuring patient confidentiality and privacy. It is available through a data use agreement, underscoring our commitment to responsible data handling and usage. Our approach has been designed to minimize any potential harm and to ensure that our findings are as unbiased and fair as possible, taking into account the diverse and complex nature of medical data. We have also taken rigorous steps to ensure our research aligns with these values.

\section*{Acknowledgments}

This work is supported by the National Key R\&D Program of China (No. 2022ZD0116401).

\clearpage
\newpage

\bibliographystyle{named}
\bibliography{ref}

\clearpage
\newpage
\appendix
\onecolumn

\input{sections/appendix}

\end{document}

%% file: sections/abstract.tex
\begin{abstract}

The integration of multimodal Electronic Health Records (EHR) data has significantly improved clinical predictive capabilities. Leveraging clinical notes and multivariate time-series EHR, existing models often lack the medical context relevent to clinical tasks, prompting the incorporation of external knowledge, particularly from the knowledge graph (KG). Previous approaches with KG knowledge have primarily focused on structured knowledge extraction, neglecting unstructured data modalities and semantic high dimensional medical knowledge. In response, we propose REALM, a Retrieval-Augmented Generation (RAG) driven framework to enhance multimodal EHR representations that address these limitations. Firstly, we apply Large Language Model (LLM) to encode long context clinical notes and GRU model to encode time-series EHR data. Secondly, we prompt LLM to extract task-relevant medical entities and match entities in  professionally labeled external knowledge graph (PrimeKG) with corresponding medical knowledge. By matching and aligning with clinical standards, our framework eliminates hallucinations and ensures consistency. Lastly, we propose an adaptive multimodal fusion network to integrate extracted knowledge with multimodal EHR data. Our extensive experiments on MIMIC-III mortality and readmission tasks showcase the superior performance of our REALM framework over baselines, emphasizing the effectiveness of each module. REALM framework contributes to refining the use of multimodal EHR data in healthcare and bridging the gap with nuanced medical context essential for informed clinical predictions.

\end{abstract}

%% file: sections/1.intro.tex
\section{Introduction}

The advent of Electronic Health Records (EHR) marks a pivotal advancement in the way patient data is gathered and analyzed, contributing to a more effective and informed healthcare delivery system for clinical prediction~\cite{ma2023aicare,gao2024comprehensive,zhu2024prism,zhang2024domaininvariant,liao2024learnable}. This advancement is largely attributed to the utilization of multimodal EHR data, which primarily includes clinical notes and multivariate time-series data from patient records~\cite{zhang2022m3care,wang2024recentEHRsurvey,zhang2023improving}. Such data types are integral to healthcare prediction tasks, mirroring the holistic approach practitioners adopt by leveraging various patient data points to inform their clinical decisions and treatment strategies, rather than depending on a single data source~\cite{xyx2023vecocare}. Deep learning-based methods have become the mainstream approach, processing multimodal data to learn a mapping from heterogeneous inputs to output labels~\cite{choi2017gram,ma2018kame,zhang2022m3care}. However, in contrast to healthcare professionals, who have a deep understanding of medical contexts through extensive experience and knowledge, neural networks trained from scratch lack these insights into medical concepts~\cite{miotto2018deep}. Without deliberate integration of external knowledge, these networks often lack the ability or sensitivity to recognize crucial disease entities or laboratory test results within the EHR, essential for accurate prediction tasks~\cite{zhu2024prompting}. In response, some recent studies have begun incorporating knowledge graphs to infuse additional medical insights into their analyses~\cite{ye2021medpath,gao2022medml}. These graphs offer a supplementary layer of clinically relevant concepts, thereby enhancing the model's ability to provide contextually meaningful representations and interpretable evidence~\cite{yang2023kerprint}. Despite these advancements, formidable limitations remain, underscoring the imperative need for continuous research in integrating insights from knowledge graphs to refine and enhance the use of multimodal EHR data in healthcare.

Previous methods integrating external medical knowledge into EHR data analysis have focused on mining hierarchical and structured knowledge from clinical-context knowledge graphs. However, these approaches primarily extract medical concepts—entity names and their relationships into a graph—with limited direct contribution to predictive tasks (\textbf{Limitation 1}). Furthermore, they tend to extract entities only from structured data modalities, such as ICD disease codes, patient conditions, procedures, and drugs, neglecting the unstructured modalities. Although extracting knowledge from unstructured data is more challenging, both clinical notes and time-series modalities are more common and practical~\cite{rajkomar2018scalable} (\textbf{Limitation 2}). With Large Language Models (LLMs) like GPT-4~\cite{openai2023gpt4} demonstrating strong capabilities in diverse clinical tasks~\cite{zhu2024prompting,wornow2023shaky} and serving as large medical knowledge graphs (KGs)~\cite{sun2023head}, it is feasible to use LLM to bridge complex medical knowledge from KGs with multimodal EHR data. By prompting the LLM, GraphCare~\cite{jiang2023graphcare} constructs a GPT-KG on structured condition, procedure, and drug record data, with triples (entity 1, relationship, entity 2) and further employs graph neural networks for downstream tasks. This paradigm, however, faces the hallucination issue of LLMs, where incorrect or fabricated information may arise~\cite{zhang2023hallucinationInLLM} (\textbf{Limitation 3}).

To overcome these limitations, we propose utilizing LLM in a Retrieval-augmented Generation (RAG) approach~\cite{lewis2020rag}. The RAG process links the unstructured modalities and external KG with LLM's semantic reasoning capabilities~\cite{wang2023can}. Despite its apparent simplicity, applying this intuition to clinical tasks presents technical challenges:

\textbf{Challenge 1: How to extract entities from multimodal EHR data and match these entities with external KG consistently?} Extracting entities from the diverse and complex formats of EHR data (including clinical notes and multivariate time-series data) presents a significant challenge. Moreover, unlike structured codes where it can directly compare the code-related entities' embedding with KG's entity, the entities extracted by LLM using prompts have hallucination issues. Accurately matching extracted entities with those in an external knowledge graph while eliminating the potential for hallucinations posed by LLMs is crucial for maintaining the integrity and reliability of the clinical prediction tasks~\cite{imrie2023redefining}.

\textbf{Challenge 2: How to encode and incorporate retrieved knowledge with original data modalities?} The extracted textual knowledge should be encoded using a sentence-level embedding model, thus posing a challenge in the selection of long-context supported models~\cite{xiao2023efficient}. Additionally, effectively incorporating retrieved knowledge with the original data modalities to enhance prediction accuracy without compromising interpretability~\cite{ye2021medpath} or introducing additional complexity into the model is vital as well.

To these ends, We propose REALM framework tackling the above limitations and challenges with the following approaches, which are our three-fold contributions:

\begin{itemize}[leftmargin=*]
    \item We design the RAG-driven multimodal enhancement framework for clinical notes and time-series EHR data (\textbf{Response to Limitation 1}). REALM leverages the capabilities of LLMs and professionally labeled external large medical knowledge graphs. We retrieve the medical entities by prompting LLM, match them in KG with detailed checking and alignment to ensure no hallucination (\textbf{Response to Limitation 3}). Apart from simple triples of entities, we also include much more knowledge and their relationships by extending the entities' definition and description and encode the long context medical knowledge into LLM embedding, which allows for capturing more complex semantic medical background knowledge that contains task-relevant insights (\textbf{Response to Limitation 2}).
    \item Methodologically, our RAG-driven entity extraction and matching process stands with a clinical standard that all knowledge comes from the professional medical knowledge graph (PrimeKG) with hallucination elimination and consistency guarantees. By carefully comparing LLM-generated entities with original data and employing threshold-based retrieval and review processes, we align the knowledge with external KGs. By extending the definition and description of entities beyond simple triples, REALM captures more complex semantic medical background knowledge. The overall process is designed to prevent hallucinations and preserve the high-level medical context from knowledge bases, ensuring the reliability of the extracted information and allowing for the inclusion of a broader range of knowledge and relationships (\textbf{Response to Challenge 1}). To infuse the extracted knowledge and with consideration of heterogeneity, we design an adaptive multimodal fusion network with self-attention and cross-attention mechanism that attentively learns each modality and fuses the final representation for downstream tasks (\textbf{Response to Challenge 2}).
    \item Our extensive experiments demonstrate REALM's superior performance on MIMIC-III mortality and readmission tasks and its effectiveness of our designed each module. Additionally, to meet the practical requirements of clinical use, we conduct an evaluation on model robustness to less training samples showing REALM's remarkable resilience against data sparsity. Moreover, the evaluation on quality of retrieved entities reflects the soundness of retrieved medical entities. Importantly, all operations in our REALM framework  are conducted offline, ensuring privacy and data security.
\end{itemize}

%% file: sections/2.related_work.tex
\section{Related Work}

\subsection{Multimodal EHR Learning}

The evolution of medical technology has enabled the analysis of various medical modalities—ranging from clinical notes and time-series laboratory test data to demographics, conditions, procedures, drugs, and medical imaging. Noteworthy efforts in multimodal learning for healthcare include
% the RAIM framework~\cite{xu2018raim}, which integrates continuous monitoring data (e.g., ECG, heart rate) with discrete clinical events (e.g., interventions, lab tests) for patient decompensation prediction. Similarly, ~\citet{gao2020compose} have developed a model that encodes both trial criteria text and patient EHR data to facilitate patient-trial matching. Diverse fusion architectures for multimodal data have been explored, such as those by ~\citet{huang2020multimodal} for Pulmonary Embolism classification and by ~\citet{lee2019predicting}, who combined neuroimaging biomarkers with longitudinal data to predict Alzheimer's disease progression. Moreover, 
M3Care~\cite{zhang2022m3care} which compensates for the missing modalities by imputing task-related information in the latent space through auxiliary information from similar patients. M3Care leverages a task-guided modality-adaptive similarity metric to effectively handle missing modalities without relying on unstable generative models. The work of ~\citet{zhang2023improving} further explored the irregularity of time intervals in time-series EHR data and clinical notes via a time attention mechanism. Notably, ~\citet{xyx2023vecocare} introduced an innovative approach for joint learning from visit sequences and clinical notes, employing Gromov-Wasserstein Distance for contrastive learning and dual-channel retrieval to enhance patient similarity analysis. ~\citet{lee2023learning} proposed a unified framework for learning across all EHR modalities, eschewing separate imputation modules in favor of modality-aware attention mechanisms.

Although the methods mentioned above perform well across multiple joint modalities, a common drawback is their limited consideration of incorporating clinical background information, wherein external medical knowledge could provide significant insights into the EHR data. Furthermore, the absence of semantic medical knowledge renders the training-from-scratch pipeline more challenging to converge, especially when data is scarce in practical clinical settings.

\subsection{Incorporating External Knowledge for EHR}

Addressing the need to blend clinical background knowledge with EHR data, numerous studies have leveraged medical knowledge graphs (KGs) to enhance the EHR data representation learning process, thereby augmenting predictive performance. Techniques such as utilizing the ancestor information of nodes within KGs have been employed to refine medical representation learning, as seen in GRAM~\cite{choi2017gram}, which integrates hierarchical medical ontologies via a graph attention network. KAME~\cite{ma2018kame} builds on this by embedding ontology information throughout the prediction process, enriching the contextual understanding of models.
% Other approaches, like MedPath~\cite{ye2021medpath}, utilize graph neural networks to capture and incorporate high-order connections from KGs into input representations, enhancing the relevancy and utility of external knowledge. 
Collaborative graph learning models, such as CGL~\cite{ijcai2021CGL}, explore patient-disease interactions and domain knowledge, while KerPrint~\cite{yang2023kerprint} focus on addressing knowledge decay on multiple time visits. The advent of Large Language Models (LLMs) as comprehensive knowledge bases~\cite{sun2023head} offers new possibilities, exemplified by GraphCare~\cite{jiang2023graphcare}, which creates a KG from structured EHR data for GNN learning, though it faces challenges related to content hallucination.

These efforts predominantly concentrate on extracting knowledge from structured medical data, overlooking the rich semantic information embedded in unstructured EHR data. This oversight limits the potential for fully leveraging the depth of knowledge contained within EHRs, highlighting the need for methodologies that encompass both structured and unstructured data modalities.

%% file: sections/3.problem_formulation.tex
\section{Problem Formulation}

The electronic health records (EHR) dataset encompasses both structured and unstructured data, represented respectively by multivariate time-series data and clinical notes. For the purpose of analysis, these two modalities are initially treated independently to derive embeddings from the raw data matrix $\bm{X}$. Specifically, multivariate time-series data, denoted as $\bm{X_{TS}} \in \mathbb{R}^{T \times F}$, is characterized by $T$ visits and $F$ numeric features. Concurrently, clinical notes, represented as $\bm{X}_{Text}$, recorded at each patient visit. Accompanying these modalities, the temporal information is encapsulated in the timestamp vector $\bm{X}_{Time} \in \mathbb{R}^{T}$, with $T$ signifying the respective visit times.

Furthermore, external knowledge graphs (KGs) are introduced to enhance the personalized representation of each individual patient. Information in the KG serve as a supplemental knowledge base reference.

The prediction objective is conceptualized as a binary classification task, encompassing the prediction of in-hospital mortality and readmission. Leveraging the comprehensive patient information derived from EHR data and KG, the model endeavors to predict specific clinical tasks. Formally, the prediction task is articulated as $\hat{y} = \text{Framework}(\bm{X}_{TS}, \bm{X}_{Text}, \bm{X}_{Time}, KG)$, where $\hat{y}$ represents the specific prediction label. Such formulation establishes a comprehensive framework for predicting clinical outcomes by amalgamating diverse data modalities and external knowledge sources.

The notations and their descriptions in the paper are shown in Table~\ref{tab:notation}. 

\begin{table}[htbp]
\footnotesize
\centering
\caption{\textit{Notations symbols and their descriptions}}
\label{tab:notation}
\begin{tabular}{c|p{5.5cm}}
\toprule
Notations & Descriptions \\
\midrule
$N$                                               & Number of patients\\
 $KG$&External knowledge graphs\\
$\bm{X}_{TS} $                                         & Time series data of one patient\\
$ \bm{X}_{Text} $                                      & Clinical records of one patient\\
$ \bm{X}_{Time} $                                      & Record time for certain modality of one patient\\
$ \bm{X}_{RAG}$                                        & Retrieved texts relative to time series or text data of one patient\\
$\bm{X}_{Time}$                                        & Visiting timestamps of one patient\\
$T$                                               & Number of visits for a certain patient\\
$D$                                               & Embedding dimension of a single modal\\
$F$                                               & Number of features in time series\\
\midrule
$\bm{h}_{i} \in \mathbb{R}^{T \times d}, \bm{h} $ & Representation of a single modality, fused to representation \( \bm{h} \), \(d\) is each modal's embedding dimension \\
$E_{TS}$                                          & Extracted entity set for one time series EHR data\\
$E_{Text}$                                        & Extracted entity set for one clinical notes data\\
$\theta$                                          & Cosine similarity of two Embedding vectors\\
$\epsilon$                                        & Threshold for selecting anomalies in time-series data\\
$\eta$                                            & Threshold for matching  extracted entities with nodes in knowledge graph\\
$s_i$                                             & z-score of i-th feature of one patient\\
\( \bm{z} \)                                      & Final representation of a patient\\
\bottomrule
\end{tabular}

\end{table}

%% file: sections/4.methods.tex
\section{Methodology}

\begin{figure*}[htbp]
  \centering
  \includegraphics[width=0.8\linewidth]{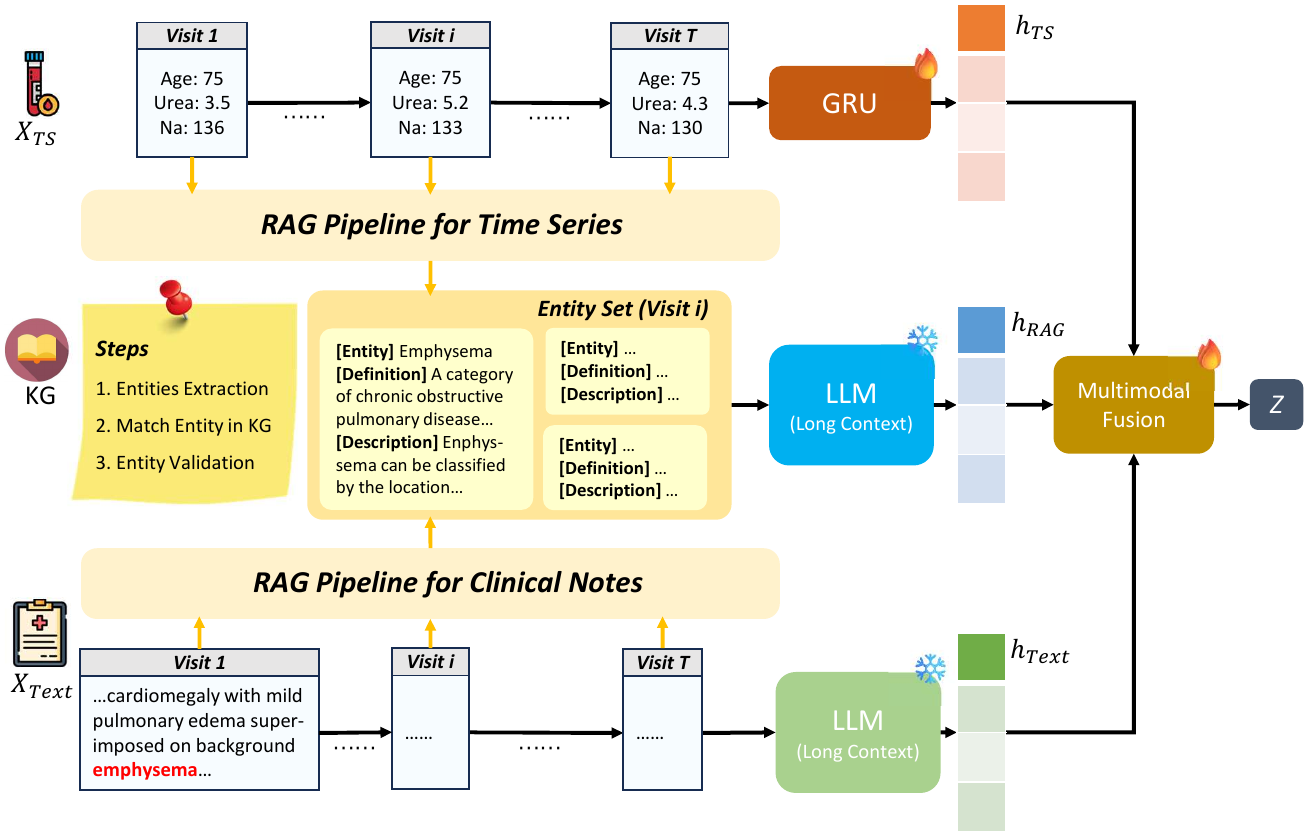}
  \caption{\textit{Overall architecture of our proposed REALM framework.}}
  \label{fig:main_fig}
\end{figure*}

\subsection{Overview}

Figure~\ref{fig:main_fig} shows the overall framework architecture of our proposed REALM model. It consists of three main modules:
\begin{itemize}[leftmargin=*]
    \item \textbf{Multimodal EHR Embedding Extraction } applies GRU as embedding model for time series $\bm{X}_{TS}$ and LLM for text records $\bm{X}_{Text}$ , which supports for long context inference at once. Readable data are transferred into embeddings $\bm{h}_{TS}$ and $\bm{h}_{Text}$.
    \item \textbf{RAG-Driven Enhancement Pipeline} retrieves relevant knowledge raw input. We design a rule-based algorithm to find outlier features from time series $\bm{X}_{TS}$, and optimize LLM prompts to extract disease entities from clinical notes $\bm{X}_{Text}$. Then semantic based retrieval match extracted entities to relative nodes from KGs over threshold $\epsilon$ or $\eta$. After that we get external information $\bm{X}_{RAG}$, and encode them into $\bm{h}_{RAG}$ respectively.
    \item \textbf{Multimodal Fusion Network} gets embedding $\bm{h}_{i}$ from input modality $\bm{X}_{i}$ and fuses them in an adaptive way to get an enhances representation $\bm{z}$.
\end{itemize}

\subsection{Multimodal EHR Embedding Extraction}

We delve into the techniques used for embedding extraction from multimodal EHR, emphasizing the transformation from raw, human-readable inputs $\bm{X}$ to deep semantic embeddings $\bm{h}$ for a thorough analysis guided by the enhanced RAG.

When dealing with time-series data $X_{TS}$, we employ the Gated Recurrent Unit (GRU) network. GRU is a highly efficient variant of recurrent neural networks, capable of capturing the time dependencies in sequence data and encoding this time-related information into $h_{TS}$, the output from the time series encoder. We choose GRU due to its exceptional ability to model time in long sequence data and its potential to tackle long-term dependencies.
\begin{equation}
\bm{h}_{TS} = {\rm GRU}(X_{TS})
\end{equation}

For text records $\bm{X}_{Text}$, we incorporated a LLM encoder to obtain text embeddings $\bm{h}_{Text}$. The primary reason for choosing LLM as the heart of the text encoder is its outstanding capability to handle long text contexts. Although the BERT model excels in numerous natural language processing tasks, its maximum input length of 512 tokens can be a limitation, potentially leading to information loss when processing long contexts. LLM encoder can handle with longer input sequences, which making it a better fit for our detailed analysis of the rich textual information in EHR.
\begin{equation}
\bm{h}_{Text} = {\rm LLM}(\bm{X}_{Text})
\end{equation}

In the realm of EHR, the temporal dimension of patient visits plays a pivotal role, with each visit characterized by a unique timestamp, denoted as $X_{Time}$. To adeptly navigate the challenges posed by the irregular and asynchronous nature of time-series data within EHR, it is essential to have an embedding strategy that can seamlessly translate these discrete temporal markers into a meaningful, continuous vector space. To this end, we draw inspiration from the advanced techniques in multi-modal EHR analysis, where time-series data is often given precedence due to its critical significance. 

Building upon the conventional Multilayer Perceptron (MLP) approach to embed time stamps $\bm{h}_{Time}$, we propose an enhanced method that leverages the sin/cos transformation, akin to the Transformer positional embedding mechanism. This approach not only captures the sequential order of visits but also preserves the cyclical continuity inherent in time-series data. By employing a sinusoidal function to encode time stamps, our model is endowed with the ability to discern the intricate inter-modality temporal relationships that are often neglected when time information is discarded. This sin/cos embedding harmonizes with the sophisticated attention mechanisms, enriching the model's capacity to prioritize relevant modalities and adapt to the dynamism of time-sensitive clinical tasks. 
% For the embeddng of time stamps($h_{Time}$), we utilized a multilayer perceptron (MLP). In scenarios involving multiple medical visits, each visit has its unique time stamp which called $X_{Time}$. To effectively incorporate this irregular time information into the model, we need a mechanism that can map these discrete time points to a continuous embedding space. Here, MLP comes into play, transforming time information into a structured vector representation that our model can understand and use.
% \yh{Learn how paper~\cite{lee2023learning} tell its story on time-series record time embedding, and also the Transformer paper. Please convert it into the sin / cos embedding (Transformer Positional Embdedding)}

\begin{equation}
\bm{h}_{Time} = {\rm MLP}(\bm{X}_{Time})
\end{equation}

By converting these three different types of data into compatible embeddings, our model lays a solid groundwork for the multimodal analysis of EHR. This strategy of embedding extraction sets the stage for further analysis tasks under the RAG framework, allowing us to accurately and comprehensively understand and analyze the complex information in EHR.

Naturally, RAG incorporates two RAG feature extraction submodules, each dealing with a different modality. These will be detailed in the following subsection.

\subsection{RAG-Driven Enhancement Pipeline}

\subsubsection{Extract Entities from Multimodal EHR Data}

To fully leverage the expert information in the knowledge graph to enhance prediction accuracy, we need to extract disease entities from time-series data and  clinical notes and match them with the information in the graph. The disease entities set in the time-series data are denoted
as $\bm{E}_{TS}$, and those in the clinical notes data can be directly denoted as $\bm{E}_{Text}$. We design two pipelines for each modality.

\begin{figure}[htbp]
    \centering
    \includegraphics[width=0.35\textwidth]{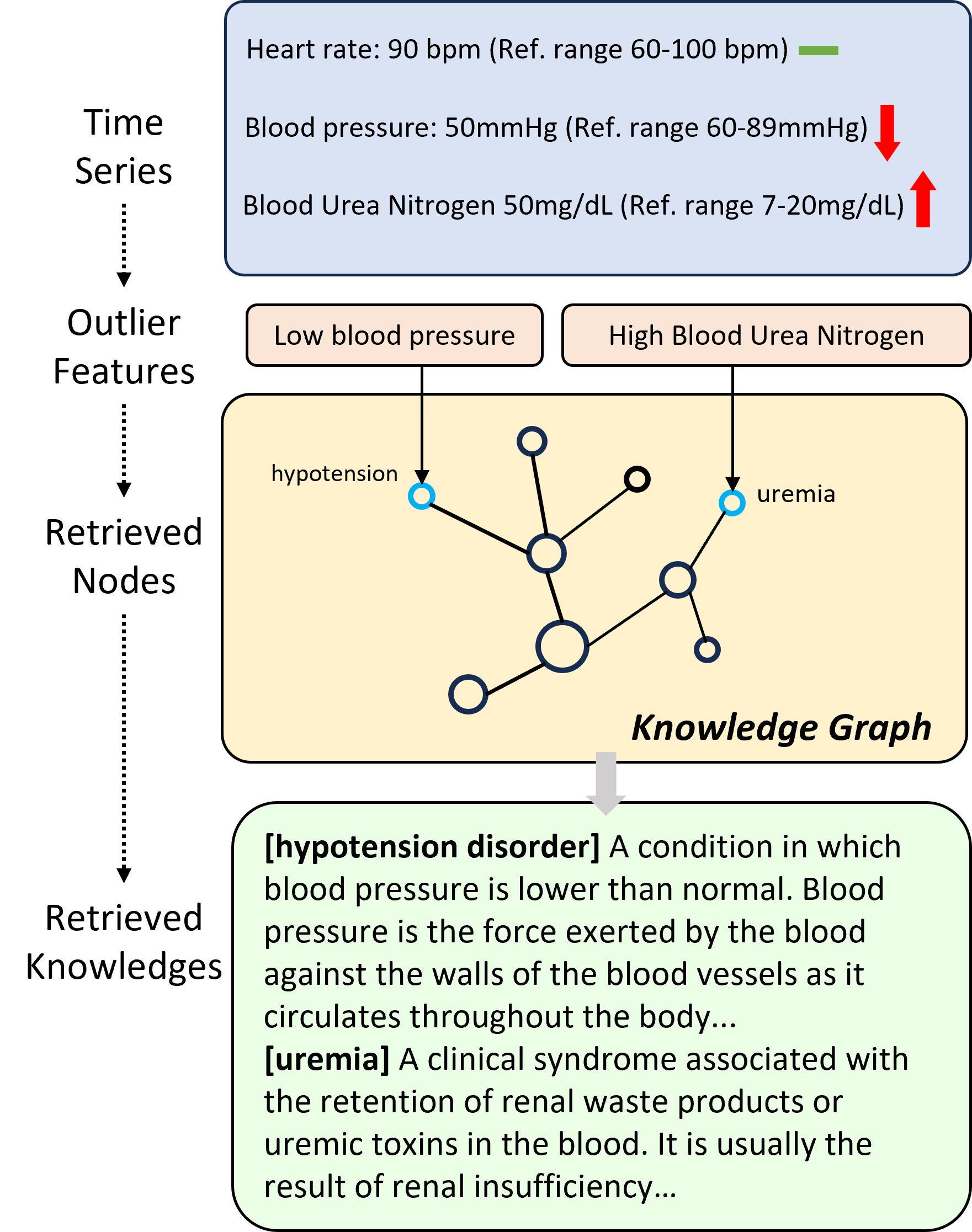}
    \caption{\textit{RAG pipeline for time series EHR modality.}}
    \label{fig:ts_rag}
\end{figure}

\paragraph{RAG module for time series.} Time series are a structured data including feature names and their values after clinical examination. Each feature name reflects parts of physical condition, which reminds us distinguishing patients through features out of reference range. As show in Figure~\ref{fig:ts_rag}, this record in total series shows a low Blood Pressure and high Blood Urea Nitrogen far beyond normal range. This reminds us the patient may suffer from hypotension and uremia. In fact, we can found these feature names in diseases defenitions and descriptions, and both lead to severe health risks. Considering continuous numeric data have obvious distribution characteristics, we can find outlier values by calculating z-score of each feature, each seemed as an entity. There are mostly more than one entity (or outlier feature) in each patient, and some are missing values, so we only focus on those not empty. For each feature $\bm{X}_{TS_i}$, we can obtain mean value and standard deviation by their reference range, and calculate z-score of each feature as below, where $s_i$ stands for z-score of $i$-th feature of one patient. 
\begin{equation}
{s_i} = \frac{\bm{X}_{TS_i}- {\rm mean}(\bm{X}_{TS_i})}{{\rm std}(\bm{X}_{TS_i})}
\end{equation}
Features over specified threshold (like 3-$\sigma$) are regarded as outlier ones, which means unhealthy physical conditions. We set $\epsilon$ as a threshold to screen out abnormal values, features with $s_i$ greater than $\epsilon$ are regarded as abnormal entities and worth extraction. 
In order to set a reasonable threshold for clinical predictions, we divided a subset manually, and observe extracted entities under different $\epsilon$, and determine one above which most entities are instructive.

\begin{figure}[htbp]
    \centering
    \includegraphics[width=0.35\textwidth]{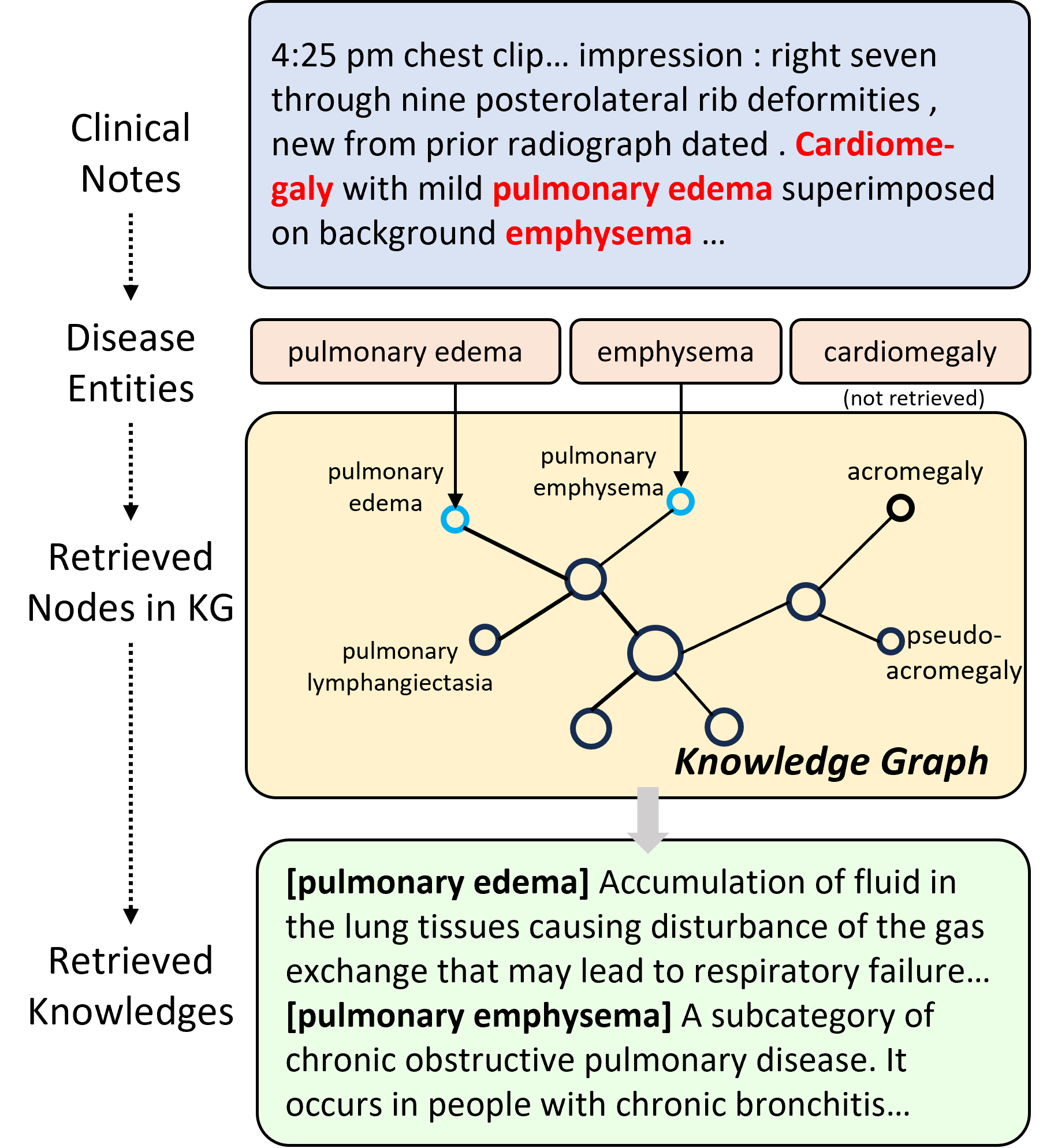}
    \caption{\textit{RAG pipeline for clinical notes modality.}}
    \label{fig:text_rag}
\end{figure}

\paragraph{RAG module for clinical text records.}Due to context limitations of models like BERT, it may cause information loss when encoding clinical notes with BERT. LLM supports for longer context, but often introduce hallucination. So we utilize LLM as entities extractor with post processing: 

\begin{enumerate}[leftmargin=*]

\item \textbf{Entities Extraction:} To reduce LLM hallucination, we use one-shot as demostration and clear instruction in prompt, guiding LLM to focus only on disease entities appearing in raw notes. When calling LLM model once, sometimes LLM may cause failure without any entities returned, so we operate in multi rounds to enlarge current extracted entity set. In $i$-th round, we concat prompt $\bm{P}_{extract}$ and clinical text notes $\bm{X}_{Text}$ together as LLM input, and we can get a set of entities in output $\bm{E}^i_{Text}$, and update total entities set with union of current one and last one.
\begin{align}
 \bm{E}^i_{Text} &= {\rm LLM}(concat(\bm{P}_{Extract},\bm{X}_{Text})) \\
 \bm{E}_{Text} &:= \bm{S}_{Text} \bigcup \bm{S}^{i}_{Text}
\end{align}
where $\bm{P}_{Extract}$  and $\bm{X}_{Text}$ represent the prompt to extract disease entities and clinical notes data respectively.

\item \textbf{Entities Refinement:} To mitigate hallucination issues of LLM, we design a post-processing procedure after extraction. This module consists of three steps: firstly, discard entities not appear in the original text; secondly, filter entities not in disease type using LLM; at last,  delete duplicated entities in semantics. After that we get a illegal entities set, and delete them from last one. This procedure may lead to new empty set, so we should loop extraction above.
\begin{equation}
\bm{E}_{Text} := \bm{E}_{Text}-\bm{E}_{illegal}
\end{equation}
We repeat step 1 and step 2 until convergence, to ensure the quality and quantity of extracted entities.

\end{enumerate}

\subsubsection{Match extracted entities with external KG}

% 基于embedding，用实体 去 匹配KG中的实体
% 阈值筛选
% 体现出此处的Check操作：

% 阈值是如何确定？过程：我们划了一个subset，根据不同的阈值，查看实体情况，人工专家校对
To accurately match the extracted entities with those in the knowledge graph, we employ dense vector retrieval based on semantics. First, we obtain embeddings of all KG nodes $Nodes$ with LLM. And we encode each entity in set $\bm{E}_{TS}$ or $\bm{E}_{Text}$ with the same LLM, to ensure embeddings align in the same vector space.
\begin{align}
    \bm{h}_{n}&= {\rm LLM}(n), n \in Nodes \\
    \bm{h}_{e}&= {\rm LLM}(e), e \in E
\end{align}

Then we use current entity $e$ as query, and compute cosine similarities between $E_e$  and all embeddings of nodes in KG $\bm{h}_n$.
\begin{equation}
\theta^{n}_{e} = \frac{{\bm{h}_n} \cdot {\bm{h}_e}}{\|{\bm{h}_n}\| \|{\bm{h}_e}\|}
\end{equation}

In our method, we set a threshold $\eta$ , to judge whether two embeddings are similar enough. If the calculated cosine similarity is greater than $\eta$, it indicates that the disease entity is closely related to this node in KGs, and we regard related attributes from this node can help us with understanding diseases meanings.

To gain an appropriate threshold, we partition a subset and examined the matching status of entities under different thresholds, followed by manual expert verification.

\subsubsection{Encode KG Knowledge}

% KG 中的 entity name, defination, description
% Prompt在最前面做拼接，经过LLM Encoder 得到RAG Embedding

To fully utilize the information of matched entities in the graph, we encode them using LLM. Firstly, we concatenate each node details together in format like (entity name, entity definition, entity description). And we join multiple node details into one sequence with specified delimiter. Then we get references knowledge as a supplement information, and get its representation with LLM, also considering of long context.
\begin{equation}
\bm{h}_{RAG} = {\rm LLM}(\bm{X}_{RAG})
\end{equation}
Additionally, when no entities found in the text, or no matched nodes in KG, we write an instruction text like "You are an experience doctor, please combine your background knowledge and patient's records to judge..." in replace of empty string instead of padding with zeros. In this way LLM can also encode an embedding containing instructive information to motivating more comprehensive understanding within the notes itself.

\subsection{Multimodal Fusion Network}

% \yh{@XSY, please implement the algs and the illustration figure}
\begin{figure}[htbp]
    \centering
    \includegraphics[width=0.8\linewidth]{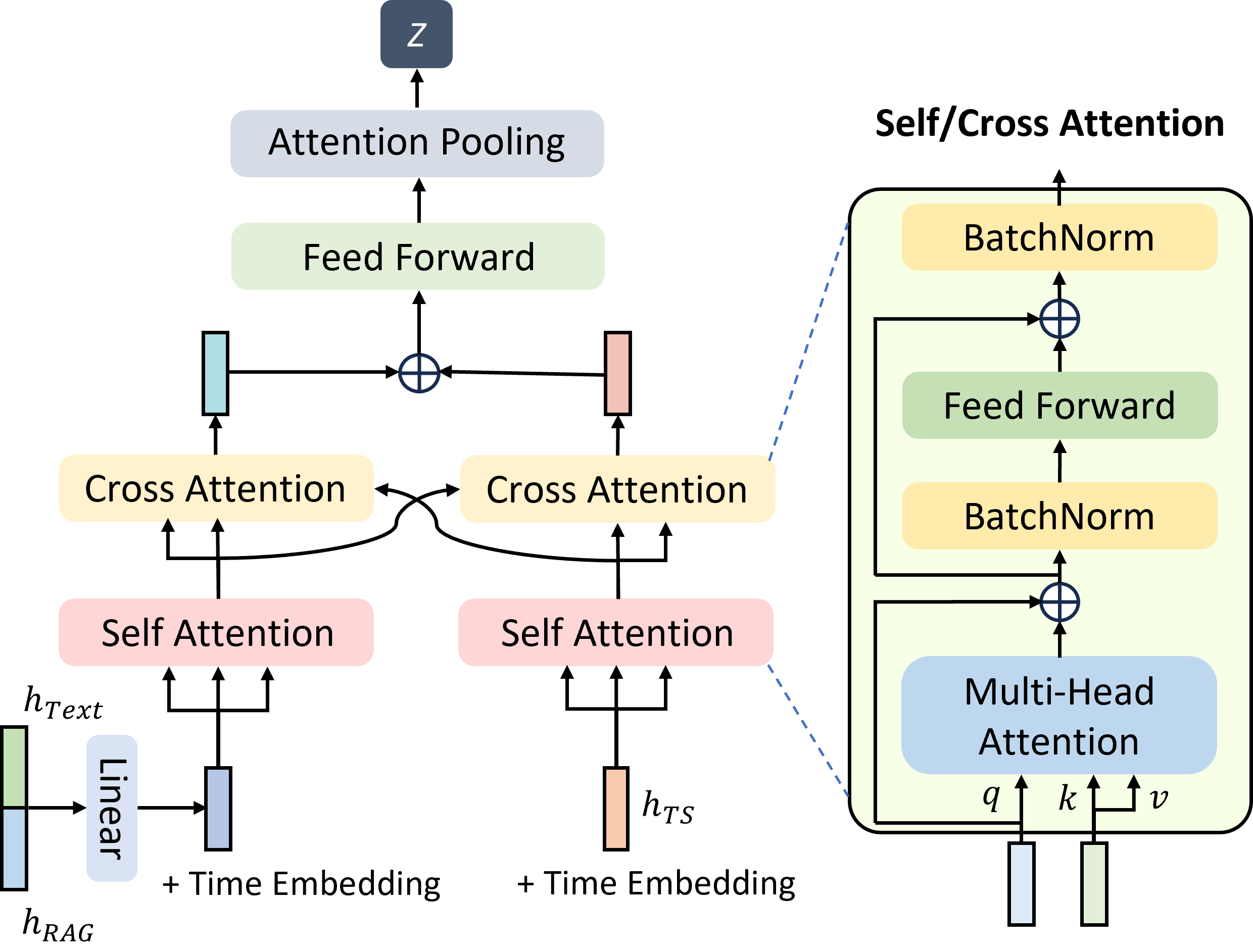}
    \caption{Fusion module. It combines multimodal embeddings with attention mechanism into a fused representation.}
    \label{fig:fusion_module}
\end{figure}

Currently, there are three learned hidden representations, denoted respectively as $\bm{h}_{TS}$, $\bm{h}_{Text}$, and $\bm{h}_{RAG}$. 
We first concatenate the hidden representations extracted from entities with those from the text, and then utilize MLP network to map them to a unified dimension.
\begin{equation}
\begin{aligned}
    \bm{h'}_{Text}={\rm MLP}({\rm Concat}(\bm{h}_{Text}, \bm{h}_{RAG}))
\end{aligned}
\end{equation}

To better integrate information from different modalities, we proposed an attention-based fusion network mainly consisting of self-attention layers and cross-attention layers. 
Specifically, we first apply self-attention to each modality. Then we use the output of one modality as the query, and the output of the other modality as the key and value to compute cross-attention.
\begin{equation}
\begin{aligned}
    \tilde{\bm{h}}_{Text}&={\rm MHSA}(\bm{h'}_{Text}+\bm{h}_{Time}),\\
    \tilde{\bm{h}}_{TS}&={\rm MHSA}(\bm{h}_{TS}+\bm{h}_{Time}), \\
    \bm{h}_{Text}&={\rm MHCA}(\tilde{\bm{h}}_{Text}, \tilde{\bm{h}}_{TS}),\\
    \bm{h}_{TS}&={\rm MHCA}(\tilde{\bm{h}}_{TS}, \tilde{\bm{h}}_{Text})
\end{aligned}
\end{equation}
where ${\rm MHSA}$ represents multi-head self-attention, ${\rm MHCA}$ represents multi-head cross-attention, and ${\bm{h}_{Time}}$ represents time embedding. In addition, we apply residual connections and BatchNorm to every multi-head attention layer and FeedForward Network.

As a result, the outputs of the two cross-attention modules have carried information from both modalities. We further sum them up and use attention poooling layer to obtain the fused information.
\begin{equation}
\begin{aligned}
    \bm{z}&=\alpha*\bm{z}_{TS}+(1-\alpha)*\bm{z}_{Text} \\
    \bm{z^*}&={\rm AttnPool}({\rm MLP}(\bm{z}))
\end{aligned}
\end{equation}
where $\alpha$ is a learnable parameter and ${\rm AttnPool}$ refers to attention pooling.

Finally, the fused representation $\bm{z^*}$ is expected to predict downstream tasks. We  pass $\bm{z^*}$ through a single-layer MLP network to obtain the final prediction results $\hat{y}$:
\begin{equation}
\begin{aligned}
    \hat{y}={\rm MLP}(\sigma(\bm{z^*}))
\end{aligned}
\end{equation}

The BCE Loss is selected as the loss function for the binary mortality outcome and readmission prediction task:
\begin{equation}
    \begin{aligned}
    \mathcal{L}(\hat{y}, y) = -\frac{1}{N}\sum_{i=1}^{N}(y_i \log(\hat{y}_i) + (1 - y_i) \log(1 - \hat{y}_i)) 
    \end{aligned}
\end{equation}
where $N$ is the number of patients within one batch, $\hat{y}\in [0,1]$ is the predicted probability and $y$ is the ground truth.

%% file: sections/5.experimental_setups.tex
\begin{table*}[!ht]
\footnotesize
\centering
\caption{\textit{In-hospital mortality and readmission prediction results on MIMIC-III.} \textbf{Bold} indicates the best performance. All metrics are multiplied by 100 for readability purposes.}
\label{tab:overall_performance_table}
\begin{tabular}{l|cccc|cccc}
\toprule
\multicolumn{1}{c|}{\multirow{2}{*}{Methods}} & \multicolumn{4}{c|}{Mortality Outcome Prediction} & \multicolumn{4}{c}{30-Day Readmission Prediction} \\
\multicolumn{1}{c|}{}    & AUROC      & AUPRC      & min(+P, Se) & F1         & AUROC      & AUPRC      & min(+P, Se) & F1    \\
\midrule
MPIM  & 85.24±1.12 & 50.52±2.56 & 50.59±2.33  & 30.53±2.33 & 78.62±1.58 & 49.30±3.01 & 49.65±2.54  & 26.61±2.20 \\
UMM  & {84.01±1.10} & {49.76±2.21} & {49.41±2.45}  & {36.21±1.90} & {77.46±1.36} & {47.81±2.55} & {47.27±1.91}  & {34.14±2.21} \\
VecoCare & {83.43±1.49} & {47.28±2.68} & {47.92±2.22}  & {42.52±2.08} & {76.93±1.82} & {46.18±2.76} & {47.22±2.63}  & {38.79±2.27} \\
M3Care & {83.33±1.24} & {47.86±2.33} & {49.96±1.99}  & {24.81±2.62} & {76.80±1.55} & {46.29±2.62} & {45.38±2.32}  & {21.51±2.23} \\ \midrule
GRAM & {84.70±1.34} & {49.21±4.45} & {49.64±2.85}  & {38.02±3.19} & {77.84±1.49} & {47.97±3.68} & {46.95±2.12}  & {35.24±2.89} \\
KAME & {84.59±1.11} & {49.48±3.37} & {49.51±2.33}  & {36.14±2.24} & {78.04±1.34} & {48.23±3.21} & {47.41±2.50}  & {31.70±2.19} \\
CGL & {84.20±1.16} & {47.64±3.47} & {47.67±2.61}  & {38.36±2.04} & {77.47±1.33} & {46.68±3.33} & {47.73±2.25}  & {35.34±2.35} \\
KerPrint & {85.29±1.21} & {51.23±3.48} & {50.88±2.24}  & {37.00±3.54} & {78.41±1.50} & {49.70±3.23} & {49.39±2.53}  & {34.31±2.35} \\ \midrule
Ours (REALM) & \textbf{86.22±0.81} & \textbf{52.64±2.47} & \textbf{50.92±2.01}  & \textbf{51.83±2.10} & \textbf{80.24±1.53} & \textbf{52.06±2.64} & \textbf{51.20±2.50}  & \textbf{50.58±2.51} \\
\bottomrule
\end{tabular}
\end{table*}

\section{Experimental Setups}

\subsection{Dataset, KG and Task Description}

Sourced from the EHRs of the Beth Israel Deaconess Medical Center, MIMIC-III dataset is extensive and widely used in healthcare research. We adhere to the benchmark pipeline~\cite{gao2024comprehensive,zhu2024pyehr} for preprocessing time-series data. 17 lab test features (include categorical features) and 2 demographic features (age and gender) are extracted. To minimize missing data, we consolidate every consecutive 12-hour segment into a single record for each patient, focusing on the first 48 records. And we follow ~\citet{khadanga2019using} to extract clinical notes. We excluded all clinical notes lacking associated chart time and removed all patients without any notes. We randomly split the dataset into training (10776 samples), validation (1539 samples) and test (3080 samples) set with 7:1:2 percentage.

The external knowledge base we utilized is PrimeKG~\cite{chandak2023buildingPrimeKG}, which integrates 20 high-quality resources to describe 17,080 diseases with 4,050,249 relationships representing ten major biological scales, including disease-associated entities. Futhermore, PrimeKG extracts textual features of disease nodes containing information about disease prevalence, symptoms, etiology, risk factors, epidemiology, clinical descriptions, management and treatment, complications, prevention, and when to seek medical attention, which are highly relevant to the clinical prediction tasks.

We conduct in-hospital mortality prediction and 30-day readmission prediction task in our experiments. Both are binary classification tasks: predicting patient mortality outcomes ($0$: alive, $1$: dead) and readmission likelihood ($0$: no readmission, $1$: possible readmission).

\subsection{Evaluation Metrics}

We adopt the following evaluation metrics, which are widely used in binary classification tasks:

\begin{itemize}[leftmargin=*]
    \item \textbf{AUROC}: This metric is our primary consideration in binary classification tasks due to its widespread use in clinical settings and its effectiveness in handling imbalanced datasets~\cite{auroc_better}.
    \item \textbf{AUPRC}: The AUPRC is particularly useful for evaluating performance in datasets with a significant imbalance between classes~\cite{kim2022auprc}.
    \item \textbf{min(+P, Se)}: This composite metric represents the minimum value between precision (+P) and sensitivity (Se), providing a balanced measure of model performance~\cite{ma2022safari}.
    \item \textbf{F1}: The F1 score is particularly useful in scenarios where an equitable trade-off between precision and recall is desired~\cite{F1}.
\end{itemize}

All these four metrics are the higher the better.

\subsection{Hyperparameters}

The batch size is consistently set at 256. For all experiments, we report performance in the form of mean±std., where we adopt bootstrap strategy for 10 times.

We conduct a grid search for the baseline models. The hyperparameters for our REALM model are: a hidden dimension of 312 and a learning rate of 6e-4.

\subsection{Baseline Models}

\paragraph{EHR Prediction Models} We include multimodal EHR baseline models (M3Care~\cite{zhang2022m3care}, MPIM~\cite{zhang2023improving}, UMM~\cite{lee2023learning}, VecoCare\cite{xyx2023vecocare}) and approaches that incorporating external knowledge from KG (GRAM~\cite{choi2017gram}, KAME~\cite{ma2018kame}, CGL~\cite{ijcai2021CGL}, KerPrint~\cite{yang2023kerprint}) as our baselines. Detailed description of each model is in Appendix.

\paragraph{Text Embedding Approaches} we compare different text embedding approaches including BERT's [CLS] token~\cite{devlin2018bert}, BGE-M3~\cite{bge-m3} and Qwen-7B's encoder~\cite{qwen}. Detailed set ups are described in Appendix.

\paragraph{Multimodal Fusion Baselines} To examine the effectiveness of our fusion network, we consider fusion methods: Add~\cite{wu2018multi}, Concat~\cite{khadanga2019using, deznabi2021predicting}, Tensor Fusion (TF)~\cite{zadeh2017tensor}, and MAG~\cite{rahman2020integrating, yang2021leverage}. Detailed description is in Appendix.

%% file: sections/6.experimental_results.tex
\section{Experimental Results}

% \yh{TODO: relative improvement}
The performance of our REALM framework on in-hospital mortality and 30-day readmission prediction tasks on the MIMIC-III dataset is summarized in Table~\ref{tab:overall_performance_table}. Our approach consistently outperforms the baseline models. Specifically, REALM achieves a significant relative improvement in AUROC (1.09\%, 2.06\%), AUPRC (2.75\%, 4.75\%), min(+P, Se) (0.79\%, 3.12\%) and F1 scores (21.90\%, 30.39\%) with the best baseline model for both tasks, indicating its superior practical applicability in real-world clinical settings.

\subsection{Ablation Studies}

\subsubsection{Comparing Each Modality with RAG-Enhancement}

To understand the contribution of RAG-enhancement to each modality, we conducted an ablation study. The results, as illustrated in Table~\ref{tab:ablation_performance_modality_fusion}, reveal that the RAG-enhanced versions of both time-series and text modalities significantly improve the model's performance. This confirms the hypothesis that enriching EHR data with external medical knowledge can effectively capture more complex semantic medical background knowledge, leading to more accurate clinical predictions.

\subsubsection{Comparing Different Fusion Network}

Our analysis extends to comparing the effectiveness of different fusion strategies for integrating time-series and text modalities. As shown in Table~\ref{tab:ablation_performance_modality_fusion}, our designed self- and cross-attention based adaptive multimodal fusion network outperforms all baseline methods in both tasks. This demonstrates the advantage of our fusion strategy in attentively learning and integrating modality-specific features for improved prediction performance. Moreover, with RAG-enhanced knowledge combining both modalities, our REALM method achieve the SOTA performance against all reduced versions.

\begin{table*}[htbp]
\footnotesize
\centering
\caption{\textit{Ablation Studies results of 1) comparing each modality with RAG-enhancement, 2) comparing different multimodal fusion network.} \textbf{Bold} indicates the best performance. All metrics are multiplied by 100 for readability purposes.}
\label{tab:ablation_performance_modality_fusion}
\begin{tabular}{l|cccc|cccc}
\toprule
\multicolumn{1}{c|}{\multirow{2}{*}{Methods}} & \multicolumn{4}{c|}{Mortality Outcome Prediction} & \multicolumn{4}{c}{30-Day Readmission Prediction} \\
\multicolumn{1}{c|}{}    & AUROC      & AUPRC      & min(+P, Se) & F1         & AUROC      & AUPRC      & min(+P, Se) & F1    \\
\midrule
TS only                  & 83.43±1.08 & 48.70±3.04 & 46.72±2.10  & 37.38±2.94 & 77.63±1.38 & 48.11±3.23 & 47.41±2.08  & 33.40±2.91 \\
TS+$RAG_{TS}$          & 84.22±0.98 & 49.80±3.15 & 48.35±1.91  & 41.10±2.95 & 78.02±1.37 & 48.36±2.98 & 47.31±2.38  & 34.39±2.73 \\
Text only                & 80.11±1.69 & 40.54±3.51 & 41.05±3.27  & 33.96±2.35 & 74.57±1.86 & 40.99±3.52 & 42.49±3.10  & 30.87±2.50 \\
Text+$RAG_{Text}$      & 81.01±1.52 & 42.92±3.43 & 42.51±3.02  & 45.13±2.44 & 74.48±1.91 & 43.38±3.70 & 43.46±3.18  & 40.01±2.91 \\ \midrule
TS+Text: \verb|Add|       & 84.72±1.03 & 48.60±3.45 & 50.05±2.59  & 46.86±2.43 & 78.23±1.74 & 48.77±3.61 & 48.76±2.87  & 47.29±2.46 \\
TS+Text: \verb|Concat|    & 85.22±0.93 & 49.94±3.14 & 49.75±1.82  & 46.51±2.18 & 78.96±1.48 & 50.08±3.27 & 50.60±2.18  & 40.61±2.02 \\
TS+Text: \verb|TF|        & 84.13±1.24 & 49.06±3.38 & 50.21±2.88  & 37.54±3.05 & 77.16±1.96 & 47.64±3.60 & 48.17±2.29  & 31.86±2.65 \\
TS+Text: \verb|MAG|       & 84.75±0.97 & 50.31±2.71 & 48.58±2.42  & 45.81±2.20 & 78.04±1.58 & 49.26±2.86 & 48.88±2.37  & 45.30±2.43 \\
TS+Text: \verb|Ours| & 85.18±0.95 & 50.68±2.64 & 47.90±2.27 & 49.81±2.37 & 78.79±1.47 & 49.69±2.92 & 48.91±2.57 & 49.94±2.36 \\
\midrule
Ours (REALM)      & \textbf{86.22±0.81} & \textbf{52.64±2.47} & \textbf{50.92±2.01}  & \textbf{51.83±2.10} & \textbf{80.24±1.53} & \textbf{52.06±2.64} & \textbf{51.20±2.50}  & \textbf{50.58±2.51} \\
\bottomrule
\end{tabular}
\end{table*}

\subsubsection{Comparing Text Embedding Models}

\begin{table}[!ht]
\footnotesize
\centering
\caption{\textit{Ablation study results of suing different text embedding models.} \textbf{Bold} indicates the best performance. All metrics are multiplied by 100 for readability purposes.}
\label{tab:ablation_performance_emb_model}
% \resizebox{\linewidth}{!}{
\begin{tabular}{cc|ccc}
\toprule
Tasks & Metrics                & BERT           & BGE-M3         & Qwen-7B  \\ \hline
\multirow{4}{*}{Out.}     & AUROC       & 83.66±1.34     &  84.72±0.97    & \textbf{86.22±0.81}  \\
                             & AUPRC       & 48.22±3.13     &  50.42±2.88    & \textbf{52.64±2.47}  \\
                             & min(+P, Se) & 48.39±3.26     &  49.29±2.55    & \textbf{50.92±2.01}       \\
                             & F1          & 43.46±2.61     &  49.40±2.41    & \textbf{51.83±2.10}       \\ \midrule
\multirow{4}{*}{Read.} & AUROC       & 76.55±1.89     &  78.03±1.63    & \textbf{80.24±1.53}       \\
                             & AUPRC       & 46.10±3.17     &  49.10±3.28    & \textbf{52.06±2.64}       \\
                             & min(+P, Se) & 46.10±3.10     &  47.67±2.41    & \textbf{51.20±2.50}       \\
                             & F1          & 39.68±2.73     &  48.81±2.22    & \textbf{50.58±2.51}       \\
\bottomrule
\end{tabular}
% }
\end{table}

The impact of using different text embedding models on the performance of our REALM framework is also explored. Table~\ref{tab:ablation_performance_emb_model} highlights that the Qwen-7B model, with its extensive training data and long-context support, significantly outperforms BERT and BGE-M3 in all metrics. This suggests that leveraging advanced large language models for embedding clinical notes can enhance the model's understanding of complex medical narratives.

\subsection{Further Analysis}

\subsubsection{Robustness to Data Sparsity}

\begin{figure}[!ht]
    \centering
    \includegraphics[width=0.8\linewidth]{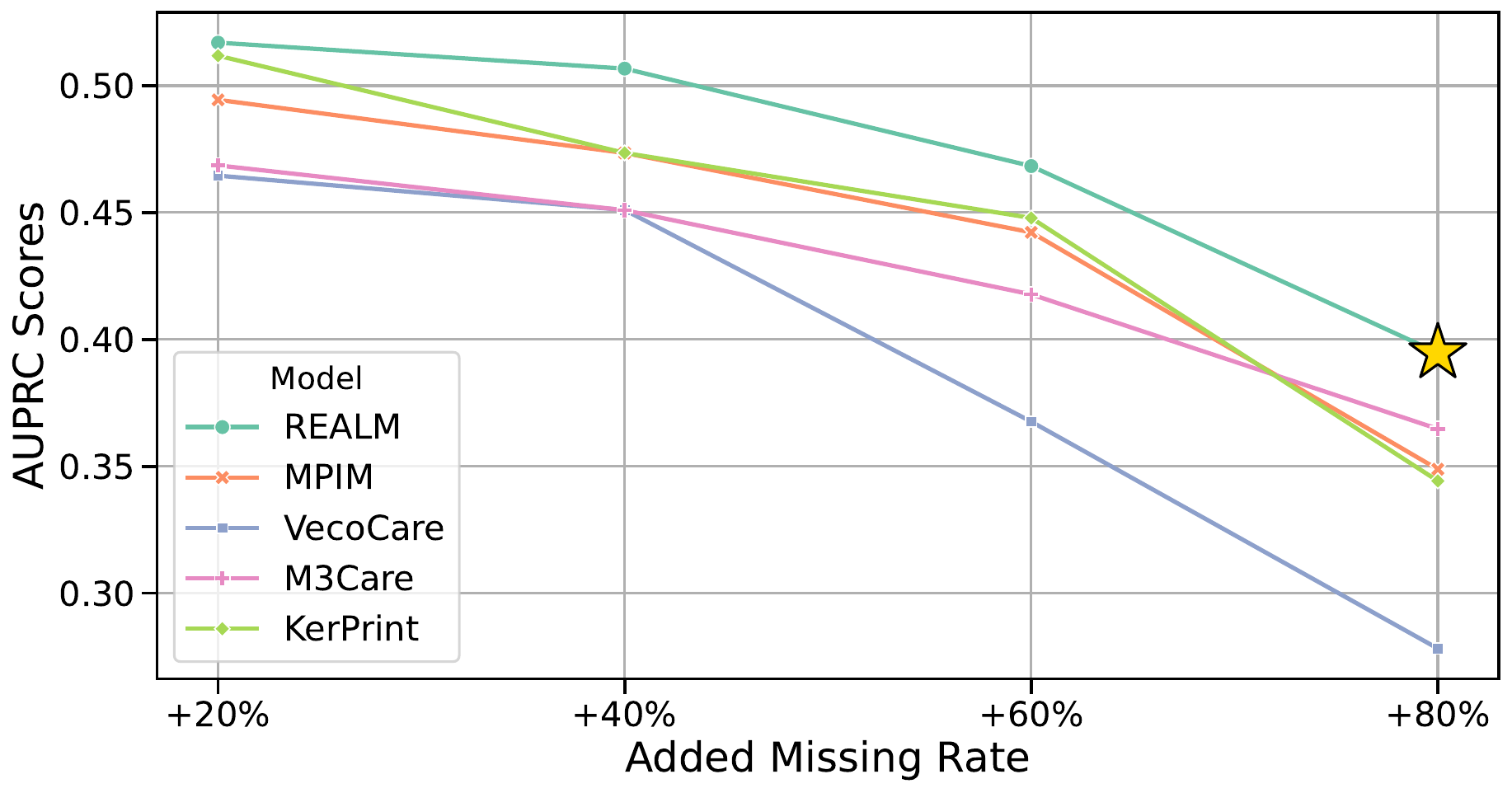}
    \caption{\textit{AUPRC Performance across 4 Sparsity Levels on MIMIC-III mortality outcome task.} REALM exhibits better performance on multiple missing rate levels than recent SOTA baselines.}
    \label{fig:missing_rates}
\end{figure}

To evaluate the robustness of our REALM framework against data sparsity, we conducted experiments by artificially reducing the dataset's completeness by 20\%, 40\%, 60\% and 80\%. As depicted in Figure~\ref{fig:missing_rates}, REALM demonstrates remarkable resilience, outperforming other recent SOTA models even under extreme data scarcity. This robustness is particularly crucial in clinical environments where large amount of data collection is often challenging, making REALM a valuable tool for real-world applications.

\subsubsection{Evaluation of Quality of Retrieved Entities}

\begin{figure}[!ht]
    \centering
    \includegraphics[width=0.8\linewidth]{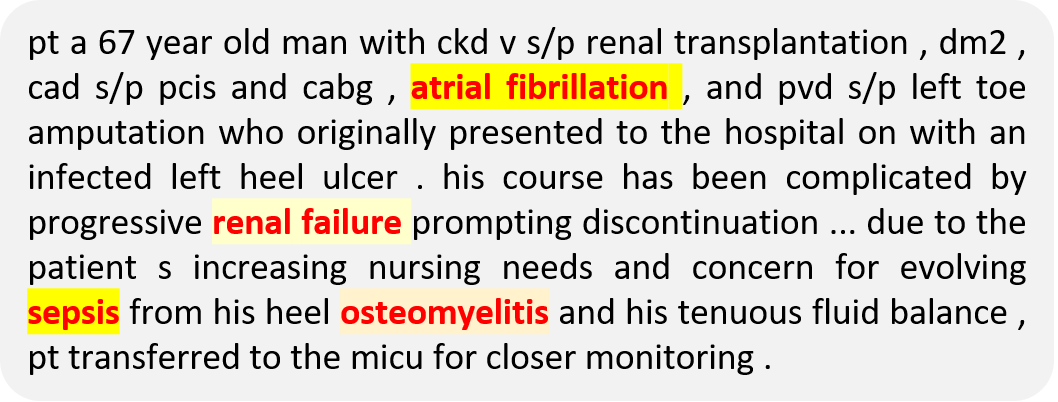}
    \caption{\textit{Case study of retrieved entities in original clinical notes with importance score computed.} The deeper yellow background color denotes higher importance score.}
    \label{fig:xgb_eval}
\end{figure}

We take the entities extracted by RAG pipeline as input to XGBoost model to calculate the importance of the entities, thereby indirectly measuring the contribution of the entities to the prediction task. Figure~\ref{fig:xgb_eval} shows the medical record of a patient's visit and the extracted disease type entities, among which "atrial fibrillation" and "sepsis" have the highest importance scores, followed by "osteomyelitis", and finally "renal failure". By examining the nodes and attribute information corresponding to each disease entity in the knowledge graph, we find that they are all relatively dangerous diseases in clinical practice, and patients have a high probability of experiencing mortality. This reflects the effectiveness of our proposed RAG-driven process.

%% file: sections/7.discussions.tex
% \section{Discussions}

% \subsection{Limitations}

% \subsection{Future Work}

%% file: sections/8.conclusions.tex
\section{Conclusions}

In this work, we propose REALM, a RAG-driven multimodal EHR data representation learning framework that incorporates time-series EHR, clinical notes data and external knowledge graph for healthcare prediction. REALM framework comprehensively leverages LLM's semantic reasoning ability, long context encoding capacity, and knowledge graph's medical context. REALM framework achieve SOTA performance on MIMIC-III datasets' in-hospital mortality and 30-day readmission tasks, showcasing its effectiveness of incorporating knowledge from external knowledge bases. Our work marks a step towards more effective utilization of EHR data in healthcare, offering a potent solution to enhance clinical representations with external knowledge and LLMs.

%% file: sections/appendix.tex
\section{Baseline Models Description Detail}

\subsection{EHR Prediction Models}

We include multimodal EHR baseline models (M3Care~\cite{zhang2022m3care}, MPIM~\cite{zhang2023improving}, UMM~\cite{lee2023learning}, VecoCare\cite{xyx2023vecocare}) and approaches that incorporating external knowledge from KG (GRAM~\cite{choi2017gram}, KAME~\cite{ma2018kame}, CGL~\cite{ijcai2021CGL}, KerPrint~\cite{yang2023kerprint}) as our baselines.

\begin{itemize}[leftmargin=*]
%%%%%%%%% MultiModal
    \item M3Care~\cite{zhang2022m3care} (KDD-2022) proposes an end-to-end model to handle missing modalities in multimodal healthcare data by imputing task-related information in the latent space using a task-guided modality-adaptive similarity metric.
    \item MPIM~\cite{zhang2023improving} (ICML-2023) models the irregularity of time intervals in time-series EHR data and clinical notes via gating mechanism and apply interleaved attention mechanism for modality fusion.
    \item UMM~\cite{lee2023learning} (MLHC-2023) introduces Unified Multi-modal Set Embedding (UMSE) and Modality-Aware Attention (MAA) with Skip Bottleneck (SB) to handle challenges in data embedding and missing modalities in multi-modal EHR.
    \item VecoCare~\cite{xyx2023vecocare} (IJCAI-2023) addresses the challenges of synthesizing information from structured and unstructured EHR data using a Gromov-Wasserstein Distance-based contrastive learning and an adaptive masked language model.
%%%%%%%%% With KG
    \item GRAM~\cite{choi2017gram} (KDD-2017) enhances EHRs with medical ontologies, using an attention mechanism to represent medical concepts by their ancestors in the ontology for improved predictive performance and interpretability.
    \item KAME~\cite{ma2018kame} (CIKM-2018) employs a knowledge attention mechanism to learn embeddings for nodes in a knowledge graph, improving the accuracy and robustness of health information prediction while providing interpretable disease representations.
    \item CGL~\cite{ijcai2021CGL} (IJCAI-2021) proposes a collaborative graph learning model to explore patient-disease interactions and incorporate medical domain knowledge, integrating unstructured text data for accurate and explainable health event predictions.
    \item KerPrint~\cite{yang2023kerprint} (AAAI-2023) offers retrospective and prospective interpretations of diagnosis predictions through a time-aware KG attention method and an element-wise attention method for selecting candidate global knowledge, enhancing interpretability.
    % \item GraphCare~\cite{jiang2023graphcare} (ICLR-2024) generates patient-specific knowledge graphs from external KGs and LLMs to improve healthcare predictions with a Bi-attention Augmented (BAT) GNN, demonstrating significant improvements in various healthcare prediction tasks.
\end{itemize}

\subsection{Text Embedding Approaches}

We compare different text embedding approaches including BERT's [CLS] token~\cite{devlin2018bert}, BGE-M3~\cite{bge-m3} and Qwen-7B's encoder~\cite{qwen}.

\begin{itemize}[leftmargin=*]
    \item BERT~\cite{devlin2018bert}: It excels in understanding context and nuance across a wide range of tasks. We use the [CLS] token from the pretrained BERT-base model for text embedding. We clip the clinical notes if they exceed the maximum token size limit of 512. The hidden dimension of BERT is 768.
    \item BGE-M3~\cite{bge-m3}: Targets multi-language and cross-language text vectorization using a vast, diverse dataset. It is specifically designed for feature extraction and retrieval, accepting up to 8192 tokens with a hidden dimension of 1024.
    \item Qwen-7B~\cite{qwen}: A large language model pretrained on 3TB of data, supporting extensive task adaptability. It can handle contexts up to 8192 tokens, making it capable of processing patients' clinical notes records.
\end{itemize}

\subsection{Multimodal Fusion Baselines}

To examine the effectiveness of our fusion network, we consider fusion methods: Add~\cite{wu2018multi}, Concat~\cite{khadanga2019using, deznabi2021predicting}, Tensor Fusion (TF)~\cite{zadeh2017tensor}, and MAG~\cite{rahman2020integrating, yang2021leverage}.

\begin{itemize}[leftmargin=*]
    \item Add~\cite{wu2018multi}: Simply performs element-wise addition of features from different modalities to integrate information.
    \item Concat~\cite{khadanga2019using, deznabi2021predicting}: The representations are concated to do predictions.
    \item Tensor Fusion (TF)~\cite{zadeh2017tensor}: It integrates information from multiple sources or modalities by creating a multimodal tensor representation that captures the interactions between these modalities.
    \item MAG~\cite{rahman2020integrating, yang2021leverage}: It dynamically fuse information from different modalities by adaptively weighting and integrating the features from each modality based on their relevance and contribution to the task at hand.
\end{itemize}

\section{Experimental Environment}

We train all models on a server equipped with an Nvidia RTX 3090 GPU and 64GB of RAM. The software environment includes CUDA 12.2, Python 3.11, PyTorch 2.0.1, and PyTorch Lightning 2.0.5. We use the AdamW optimizer. All models are trained over 30 epochs on patient samples from the training set, with an early stopping strategy monitored by the AUROC score on the validation set, applying a patience of 5 epochs.

\section{Prompt Details}

Here are templates when extracting entities using LLM, and combining retrieved knowledge with KG.

\begin{tcolorbox}[title=Prompt for entity extraction,
    colback=white,
    colframe=black,
    colbacktitle=white,
    coltitle=black,
    fonttitle=\bfseries,
    breakable,
    enhanced] % Adds some space between paragraphs automatically
\begin{Verbatim}[fontsize=\small]
[Instruction]

You are tasked with performing Named Entity Recognition (NER) specifically for diseases in 
a given medical case description. Follow the instructions below:

1. Input: You will receive a medical case description in the [Input].
2. NER Task: Focus on extracting the names of diseases as the target entity.
3. Output: Provide the extracted disease names in JSON format.

Ensure that the JSON output only includes the names of diseases mentioned in the provided 
[Input], excluding any additional content. The goal is to perform NER exclusively on 
disease names within the given text.

Example:

[Input]

 1:19 pm abdomen ( supine and erect ) clip  reason : sbo medical condition : 63 year old 
 woman with reason for this examination : sbo final report indication : 63-year-old woman 
 with small bowel obstruction . findings : supine and upright abdominal radiographs are 
 markedly limited due to the patient 's body habitus ...
 
[Answer]

{
    "entities": ["small bowel obstruction", "large bowel volvulus", ...]
}
\end{Verbatim}
\end{tcolorbox}

\begin{tcolorbox}[title=Complete $X_{RAG}$ demonstration with retrieved results,
    colback=white,
    colframe=black,
    colbacktitle=white,
    coltitle=black,
    fonttitle=\bfseries,
    breakable,
    enhanced]
\begin{Verbatim}[fontsize=\small]
You are an experienced doctor, and the patient's medical record includes the following 
disease entities during a visit. Below are descriptions related to these diseases. Please 
combine the severity of the patient's condition in the medical record and the severity of 
the disease to assess their health status.

References:

[Disease] acromegaly 

[Definition] Acromegaly is an acquired disorder related to excessive production of growth 
hormone (GH) and characterized by progressive somatic disfigurement (mainly involving the
face and extremities) and systemic manifestations.. A rare acquired endocrine disease 
related to excessive production of growth hormone (GH) and characterized by progressive 
somatic disfigurement (mainly involving the face and extremities) and systemic 
manifestations. 

[Description] Disease of the glandular, anterior portion of the pituitary resulting in 
hypersecretion of adenohypophyseal hormones such as growth hormone; prolactin; 
hyrotropin; luteinizing hormone; follicle stimulating hormone ; and adrenocorticotropic 
hormone. Hyperpituitarism usually is caused by a functional adenoma.

[Disease] pulmonary edema

[Definition] Accumulation of fluid in the lung tissues causing disturbance of the gas 
exchange that may lead to respiratory failure. It is caused by direct injury to the lung 
parenchyma or congestive heart failure. The symptoms may appear suddenly or gradually. 
Suddenly appearing symptoms include difficulty breathing, feeling of suffocation, and 
coughing associated with frothy sputum. Gradually appearing symptoms include difficulty 
breathing while lying in bed, shortness of breath during activity, and weight gain (in 
patients with congestive heart failure).

[Description] Excessive accumulation of extravascular fluid in the lung, an indication of 
a serious underlying disease or disorder. Pulmonary edema prevents efficient pulmonary
gas exchange in the pulmonary alveoli, and can be life-threatening.

\end{Verbatim}
\end{tcolorbox}

%% file: main.bbl
\begin{thebibliography}{}

\bibitem[\protect\citeauthoryear{Achiam \bgroup \em et al.\egroup }{2023}]{openai2023gpt4}
Josh Achiam, Steven Adler, Sandhini Agarwal, Lama Ahmad, Ilge Akkaya, Florencia~Leoni Aleman, Diogo Almeida, Janko Altenschmidt, Sam Altman, Shyamal Anadkat, et~al.
\newblock Gpt-4 technical report.
\newblock {\em arXiv preprint arXiv:2303.08774}, 2023.

\bibitem[\protect\citeauthoryear{Bai \bgroup \em et al.\egroup }{2023}]{qwen}
Jinze Bai, Shuai Bai, Yunfei Chu, Zeyu Cui, Kai Dang, Xiaodong Deng, Yang Fan, Wenbin Ge, Yu~Han, Fei Huang, et~al.
\newblock Qwen technical report.
\newblock {\em arXiv preprint arXiv:2309.16609}, 2023.

\bibitem[\protect\citeauthoryear{Chandak \bgroup \em et al.\egroup }{2023}]{chandak2023buildingPrimeKG}
Payal Chandak, Kexin Huang, and Marinka Zitnik.
\newblock Building a knowledge graph to enable precision medicine.
\newblock {\em Scientific Data}, 10(1):67, 2023.

\bibitem[\protect\citeauthoryear{Chen and Xiao}{2024}]{bge-m3}
Jianlv Chen and Shitao Xiao.
\newblock Bge m3-embedding: Multi-lingual, multi-functionality, multi-granularity text embeddings through self-knowledge distillation.
\newblock \url{https://synthical.com/article/9ffce599-0640-457c-bd1c-502cab06e8af}, 1 2024.

\bibitem[\protect\citeauthoryear{Chinchor}{1992}]{F1}
Nancy Chinchor.
\newblock Muc-4 evaluation metrics.
\newblock In {\em Proceedings of the 4th Conference on Message Understanding}, MUC4 '92, page 22–29, USA, 1992. Association for Computational Linguistics.

\bibitem[\protect\citeauthoryear{Choi \bgroup \em et al.\egroup }{2017}]{choi2017gram}
Edward Choi, Mohammad~Taha Bahadori, Le~Song, Walter~F Stewart, and Jimeng Sun.
\newblock Gram: graph-based attention model for healthcare representation learning.
\newblock In {\em Proceedings of the 23rd ACM SIGKDD international conference on knowledge discovery and data mining}, pages 787--795, 2017.

\bibitem[\protect\citeauthoryear{Devlin \bgroup \em et al.\egroup }{2018}]{devlin2018bert}
Jacob Devlin, Ming-Wei Chang, Kenton Lee, and Kristina Toutanova.
\newblock Bert: Pre-training of deep bidirectional transformers for language understanding.
\newblock {\em arXiv preprint arXiv:1810.04805}, 2018.

\bibitem[\protect\citeauthoryear{Deznabi \bgroup \em et al.\egroup }{2021}]{deznabi2021predicting}
Iman Deznabi, Mohit Iyyer, and Madalina Fiterau.
\newblock Predicting in-hospital mortality by combining clinical notes with time-series data.
\newblock In {\em Findings of the Association for Computational Linguistics: ACL-IJCNLP 2021}, pages 4026--4031, 2021.

\bibitem[\protect\citeauthoryear{Gao \bgroup \em et al.\egroup }{2022}]{gao2022medml}
Junyi Gao, Chaoqi Yang, Joerg Heintz, Scott Barrows, Elise Albers, Mary Stapel, Sara Warfield, Adam Cross, and Jimeng Sun.
\newblock Medml: fusing medical knowledge and machine learning models for early pediatric covid-19 hospitalization and severity prediction.
\newblock {\em Iscience}, 25(9), 2022.

\bibitem[\protect\citeauthoryear{Gao \bgroup \em et al.\egroup }{2024}]{gao2024comprehensive}
Junyi Gao, Yinghao Zhu, Wenqing Wang, Yasha Wang, Wen Tang, Ewen~M. Harrison, and Liantao Ma.
\newblock A comprehensive benchmark for covid-19 predictive modeling using electronic health records in intensive care, 2024.

\bibitem[\protect\citeauthoryear{Imrie \bgroup \em et al.\egroup }{2023}]{imrie2023redefining}
Fergus Imrie, Paulius Rauba, and Mihaela van~der Schaar.
\newblock Redefining digital health interfaces with large language models.
\newblock {\em arXiv preprint arXiv:2310.03560}, 2023.

\bibitem[\protect\citeauthoryear{Jiang \bgroup \em et al.\egroup }{2023}]{jiang2023graphcare}
Pengcheng Jiang, Cao Xiao, Adam Cross, and Jimeng Sun.
\newblock Graphcare: Enhancing healthcare predictions with open-world personalized knowledge graphs.
\newblock {\em arXiv preprint arXiv:2305.12788}, 2023.

\bibitem[\protect\citeauthoryear{Khadanga \bgroup \em et al.\egroup }{2019}]{khadanga2019using}
Swaraj Khadanga, Karan Aggarwal, Shafiq Joty, and Jaideep Srivastava.
\newblock Using clinical notes with time series data for icu management.
\newblock {\em arXiv preprint arXiv:1909.09702}, 2019.

\bibitem[\protect\citeauthoryear{Kim and Hwang}{2022}]{kim2022auprc}
Misuk Kim and Kyu-Baek Hwang.
\newblock An empirical evaluation of sampling methods for the classification of imbalanced data.
\newblock {\em PLoS One}, 17(7):e0271260, 2022.

\bibitem[\protect\citeauthoryear{Lee \bgroup \em et al.\egroup }{2023}]{lee2023learning}
Kwanhyung Lee, Soojeong Lee, Sangchul Hahn, Heejung Hyun, Edward Choi, Byungeun Ahn, and Joohyung Lee.
\newblock Learning missing modal electronic health records with unified multi-modal data embedding and modality-aware attention.
\newblock {\em arXiv preprint arXiv:2305.02504}, 2023.

\bibitem[\protect\citeauthoryear{Lewis \bgroup \em et al.\egroup }{2020}]{lewis2020rag}
Patrick Lewis, Ethan Perez, Aleksandra Piktus, Fabio Petroni, Vladimir Karpukhin, Naman Goyal, Heinrich K{\"u}ttler, Mike Lewis, Wen-tau Yih, Tim Rockt{\"a}schel, et~al.
\newblock Retrieval-augmented generation for knowledge-intensive nlp tasks.
\newblock {\em Advances in Neural Information Processing Systems}, 33:9459--9474, 2020.

\bibitem[\protect\citeauthoryear{Liao \bgroup \em et al.\egroup }{2024}]{liao2024learnable}
Weibin Liao, Yinghao Zhu, Zixiang Wang, Xu~Chu, Yasha Wang, and Liantao Ma.
\newblock Learnable prompt as pseudo-imputation: Reassessing the necessity of traditional ehr data imputation in downstream clinical prediction, 2024.

\bibitem[\protect\citeauthoryear{Lu \bgroup \em et al.\egroup }{2021}]{ijcai2021CGL}
Chang Lu, Chandan~K Reddy, Prithwish Chakraborty, Samantha Kleinberg, and Yue Ning.
\newblock Collaborative graph learning with auxiliary text for temporal event prediction in healthcare.
\newblock In Zhi-Hua Zhou, editor, {\em Proceedings of the Thirtieth International Joint Conference on Artificial Intelligence, {IJCAI-21}}, pages 3529--3535. International Joint Conferences on Artificial Intelligence Organization, 8 2021.
\newblock Main Track.

\bibitem[\protect\citeauthoryear{Ma \bgroup \em et al.\egroup }{2018}]{ma2018kame}
Fenglong Ma, Quanzeng You, Houping Xiao, Radha Chitta, Jing Zhou, and Jing Gao.
\newblock Kame: Knowledge-based attention model for diagnosis prediction in healthcare.
\newblock In {\em Proceedings of the 27th ACM International Conference on Information and Knowledge Management}, pages 743--752, 2018.

\bibitem[\protect\citeauthoryear{Ma \bgroup \em et al.\egroup }{2022}]{ma2022safari}
Xinyu Ma, Yasha Wang, Xu~Chu, Liantao Ma, Wen Tang, Junfeng Zhao, Ye~Yuan, and Guoren Wang.
\newblock Patient health representation learning via correlational sparse prior of medical features.
\newblock {\em IEEE Transactions on Knowledge and Data Engineering}, 2022.

\bibitem[\protect\citeauthoryear{Ma \bgroup \em et al.\egroup }{2023}]{ma2023aicare}
Liantao Ma, Chaohe Zhang, Junyi Gao, Xianfeng Jiao, Zhihao Yu, Yinghao Zhu, Tianlong Wang, Xinyu Ma, Yasha Wang, Wen Tang, Xinju Zhao, Wenjie Ruan, and Tao Wang.
\newblock Mortality prediction with adaptive feature importance recalibration for peritoneal dialysis patients.
\newblock {\em Patterns}, 4(12), 2023.

\bibitem[\protect\citeauthoryear{McDermott \bgroup \em et al.\egroup }{2024}]{auroc_better}
Matthew B.~A. McDermott, Lasse~Hyldig Hansen, Haoran Zhang, Giovanni Angelotti, and Jack Gallifant.
\newblock A closer look at auroc and auprc under class imbalance.
\newblock {\em arXiv preprint arXiv:2401.06091}, 2024.

\bibitem[\protect\citeauthoryear{Miotto \bgroup \em et al.\egroup }{2018}]{miotto2018deep}
Riccardo Miotto, Fei Wang, Shuang Wang, Xiaoqian Jiang, and Joel~T Dudley.
\newblock Deep learning for healthcare: review, opportunities and challenges.
\newblock {\em Briefings in bioinformatics}, 19(6):1236--1246, 2018.

\bibitem[\protect\citeauthoryear{Rahman \bgroup \em et al.\egroup }{2020}]{rahman2020integrating}
Wasifur Rahman, Md~Kamrul Hasan, Sangwu Lee, Amir Zadeh, Chengfeng Mao, Louis-Philippe Morency, and Ehsan Hoque.
\newblock Integrating multimodal information in large pretrained transformers.
\newblock In {\em Proceedings of the conference. Association for Computational Linguistics. Meeting}, volume 2020, page 2359. NIH Public Access, 2020.

\bibitem[\protect\citeauthoryear{Rajkomar \bgroup \em et al.\egroup }{2018}]{rajkomar2018scalable}
Alvin Rajkomar, Eyal Oren, Kai Chen, Andrew~M Dai, Nissan Hajaj, Michaela Hardt, Peter~J Liu, Xiaobing Liu, Jake Marcus, Mimi Sun, et~al.
\newblock Scalable and accurate deep learning with electronic health records.
\newblock {\em NPJ digital medicine}, 1(1):18, 2018.

\bibitem[\protect\citeauthoryear{Sun \bgroup \em et al.\egroup }{2023}]{sun2023head}
Kai Sun, Yifan~Ethan Xu, Hanwen Zha, Yue Liu, and Xin~Luna Dong.
\newblock Head-to-tail: How knowledgeable are large language models (llm)? aka will llms replace knowledge graphs?
\newblock {\em arXiv preprint arXiv:2308.10168}, 2023.

\bibitem[\protect\citeauthoryear{Wang \bgroup \em et al.\egroup }{2023}]{wang2023can}
Boshi Wang, Xiang Yue, and Huan Sun.
\newblock Can chatgpt defend its belief in truth? evaluating llm reasoning via debate.
\newblock In {\em Findings of the Association for Computational Linguistics: EMNLP 2023}, pages 11865--11881, 2023.

\bibitem[\protect\citeauthoryear{Wang \bgroup \em et al.\egroup }{2024}]{wang2024recentEHRsurvey}
Jiaqi Wang, Junyu Luo, Muchao Ye, Xiaochen Wang, Yuan Zhong, Aofei Chang, Guanjie Huang, Ziyi Yin, Cao Xiao, Jimeng Sun, and Fenglong Ma.
\newblock Recent advances in predictive modeling with electronic health records, 2024.

\bibitem[\protect\citeauthoryear{Wornow \bgroup \em et al.\egroup }{2023}]{wornow2023shaky}
Michael Wornow, Yizhe Xu, Rahul Thapa, Birju Patel, Ethan Steinberg, Scott Fleming, Michael~A Pfeffer, Jason Fries, and Nigam~H Shah.
\newblock The shaky foundations of large language models and foundation models for electronic health records.
\newblock {\em npj Digital Medicine}, 6(1):135, 2023.

\bibitem[\protect\citeauthoryear{Wu and Han}{2018}]{wu2018multi}
Aming Wu and Yahong Han.
\newblock Multi-modal circulant fusion for video-to-language and backward.
\newblock In {\em Proceedings of the Twenty-Seventh International Joint Conference on Artificial Intelligence, {IJCAI-18}}, pages 1029--1035. International Joint Conferences on Artificial Intelligence Organization, 7 2018.

\bibitem[\protect\citeauthoryear{Xiao \bgroup \em et al.\egroup }{2023}]{xiao2023efficient}
Guangxuan Xiao, Yuandong Tian, Beidi Chen, Song Han, and Mike Lewis.
\newblock Efficient streaming language models with attention sinks, 2023.

\bibitem[\protect\citeauthoryear{Xu \bgroup \em et al.\egroup }{2023}]{xyx2023vecocare}
Yongxin Xu, Kai Yang, Chaohe Zhang, Peinie Zou, Zhiyuan Wang, Hongxin Ding, Junfeng Zhao, Yasha Wang, and Bing Xie.
\newblock Vecocare: Visit sequences-clinical notes joint learning for diagnosis prediction in healthcare data.
\newblock In Edith Elkind, editor, {\em Proceedings of the Thirty-Second International Joint Conference on Artificial Intelligence, {IJCAI-23}}, pages 4921--4929. International Joint Conferences on Artificial Intelligence Organization, 8 2023.
\newblock Main Track.

\bibitem[\protect\citeauthoryear{Yang and Wu}{2021}]{yang2021leverage}
Bo~Yang and Lijun Wu.
\newblock How to leverage multimodal ehr data for better medical predictions?
\newblock {\em arXiv preprint arXiv:2110.15763}, 2021.

\bibitem[\protect\citeauthoryear{Yang \bgroup \em et al.\egroup }{2023}]{yang2023kerprint}
Kai Yang, Yongxin Xu, Peinie Zou, Hongxin Ding, Junfeng Zhao, Yasha Wang, and Bing Xie.
\newblock Kerprint: Local-global knowledge graph enhanced diagnosis prediction for retrospective and prospective interpretations.
\newblock {\em Proceedings of the AAAI Conference on Artificial Intelligence}, 37(4):5357--5365, Jun. 2023.

\bibitem[\protect\citeauthoryear{Ye \bgroup \em et al.\egroup }{2021}]{ye2021medpath}
Muchao Ye, Suhan Cui, Yaqing Wang, Junyu Luo, Cao Xiao, and Fenglong Ma.
\newblock Medpath: Augmenting health risk prediction via medical knowledge paths.
\newblock In {\em Proceedings of the Web Conference 2021}, pages 1397--1409, 2021.

\bibitem[\protect\citeauthoryear{Zadeh \bgroup \em et al.\egroup }{2017}]{zadeh2017tensor}
Amir Zadeh, Minghai Chen, Soujanya Poria, Erik Cambria, and Louis-Philippe Morency.
\newblock Tensor fusion network for multimodal sentiment analysis.
\newblock {\em arXiv preprint arXiv:1707.07250}, 2017.

\bibitem[\protect\citeauthoryear{Zhang \bgroup \em et al.\egroup }{2022}]{zhang2022m3care}
Chaohe Zhang, Xu~Chu, Liantao Ma, Yinghao Zhu, Yasha Wang, Jiangtao Wang, and Junfeng Zhao.
\newblock M3care: Learning with missing modalities in multimodal healthcare data.
\newblock In {\em Proceedings of the 28th ACM SIGKDD Conference on Knowledge Discovery and Data Mining}, KDD '22, page 2418–2428, New York, NY, USA, 2022. Association for Computing Machinery.

\bibitem[\protect\citeauthoryear{Zhang \bgroup \em et al.\egroup }{2023a}]{zhang2023improving}
Xinlu Zhang, Shiyang Li, Zhiyu Chen, Xifeng Yan, and Linda~Ruth Petzold.
\newblock Improving medical predictions by irregular multimodal electronic health records modeling.
\newblock In {\em International Conference on Machine Learning}, pages 41300--41313. PMLR, 2023.

\bibitem[\protect\citeauthoryear{Zhang \bgroup \em et al.\egroup }{2023b}]{zhang2023hallucinationInLLM}
Yue Zhang, Yafu Li, Leyang Cui, Deng Cai, Lemao Liu, Tingchen Fu, Xinting Huang, Enbo Zhao, Yu~Zhang, Yulong Chen, et~al.
\newblock Siren's song in the ai ocean: A survey on hallucination in large language models.
\newblock {\em arXiv preprint arXiv:2309.01219}, 2023.

\bibitem[\protect\citeauthoryear{Zhang \bgroup \em et al.\egroup }{2024}]{zhang2024domaininvariant}
Zhongji Zhang, Yuhang Wang, Yinghao Zhu, Xinyu Ma, Tianlong Wang, Chaohe Zhang, Yasha Wang, and Liantao Ma.
\newblock Domain-invariant clinical representation learning by bridging data distribution shift across emr datasets, 2024.

\bibitem[\protect\citeauthoryear{Zhu \bgroup \em et al.\egroup }{2024a}]{zhu2024pyehr}
Yinghao Zhu, Wenqing Wang, Junyi Gao, and Liantao Ma.
\newblock Pyehr: A predictive modeling toolkit for electronic health records.
\newblock \url{https://github.com/yhzhu99/pyehr}, 2024.

\bibitem[\protect\citeauthoryear{Zhu \bgroup \em et al.\egroup }{2024b}]{zhu2024prompting}
Yinghao Zhu, Zixiang Wang, Junyi Gao, Yuning Tong, Jingkun An, Weibin Liao, Ewen~M. Harrison, Liantao Ma, and Chengwei Pan.
\newblock Prompting large language models for zero-shot clinical prediction with structured longitudinal electronic health record data, 2024.

\bibitem[\protect\citeauthoryear{Zhu \bgroup \em et al.\egroup }{2024c}]{zhu2024prism}
Yinghao Zhu, Zixiang Wang, Long He, Shiyun Xie, Liantao Ma, and Chengwei Pan.
\newblock Prism: Leveraging prototype patient representations with feature-missing-aware calibration for ehr data sparsity mitigation, 2024.

\end{thebibliography}
